\pdfoutput=1

\documentclass{article}

\usepackage{microtype}
\usepackage{graphicx}
\usepackage{subfigure}
\usepackage{booktabs} 

\usepackage{amsmath,amssymb,amsfonts,amsthm}
\usepackage{commath}

\usepackage{hyperref}

\usepackage[accepted]{icml2020}

\icmltitlerunning{Goal-Directed Attention in DCNNs}

\begin{document}

\twocolumn[
\icmltitle{The Costs and Benefits of Goal-Directed Attention in Deep Convolutional Neural Networks}

\begin{icmlauthorlist}
\icmlauthor{Xiaoliang Luo}{ucl}
\icmlauthor{Brett D.~Roads}{ucl}
\icmlauthor{Bradley C.~Love}{ucl,ati}
\end{icmlauthorlist}

\icmlaffiliation{ucl}{Department of Experimental Psychology, University College London, London, United Kingdom.}
\icmlaffiliation{ati}{The Alan Turing Institute, London, United Kingdom}

\icmlcorrespondingauthor{Xiaoliang Luo}{xiao.luo.17@ucl.ac.uk}


\icmlkeywords{Attention, neuroscience, ICML}

\vskip 0.3in
]



\printAffiliationsAndNotice{}  

\begin{abstract}
People deploy top-down, goal-directed attention to accomplish tasks, such as finding lost keys. By tuning the visual system to relevant information sources, object recognition can become more efficient (a benefit) and more biased toward the target (a potential cost). Motivated by selective attention in categorisation models, we developed a goal-directed attention mechanism that can process naturalistic (photographic) stimuli. Our attention mechanism can be incorporated into any existing deep convolutional neural network (DCNNs). The processing stages in DCNNs have been related to ventral visual stream. In that light, our attentional mechanism incorporates top-down influences from prefrontal cortex (PFC) to support goal-directed behaviour. Akin to how attention weights in categorisation models warp representational spaces, we introduce a layer of attention weights to the mid-level of a DCNN that amplify or attenuate activity to further a goal. We evaluated the attentional mechanism using photographic stimuli, varying the attentional target. We found that increasing goal-directed attention has benefits (increasing hit rates) and costs (increasing false alarm rates). At a moderate level, attention improves sensitivity (i.e., increases $d^\prime$) at only a moderate increase in bias for tasks involving standard images, blended images, and natural adversarial images chosen to fool DCNNs. These results suggest that goal-directed attention can reconfigure general-purpose DCNNs to better suit the current task goal, much like PFC modulates activity along the ventral stream. In addition to being more parsimonious and brain consistent, the mid-level attention approach performed better than a standard machine learning approach for transfer learning, namely retraining the final network layer to accommodate the new task.
\end{abstract}


\section{Introduction}
\label{Introduction}
Imagine looking for your car keys in the kitchen. At first, one might focus on features such as small and metallic. This attention focus could lead one to false alarm to a stray fork occluded by a chopping board, but should also increase the chances of finding one's keys. To carry out this search task, the prefrontal cortex (PFC) exerts goal-directed pressure on the visual system to favour goal-relevant information \citep{Miller2001AnFunction}. Goal-directed attention can reconfigure the visual system to highlight task-relevant features and suppress irrelevant features.

Recent advances in machine learning have made it possible to investigate goal-directed attention with naturalistic stimuli, potentially paving the way for cognitive models that can be applied to richer tasks and stimulus sets \citep[cf.][]{Nosofsky2018ALearning, Peterson2018EvaluatingRepresentations,Guest2019LevelsCategorization}. Here, we extend the selective-attention mechanisms that have proven successful in cognitive models of category learning \citep{Nosofsky1986AttentionRelationship., Kruschke1992ALCOVE:Learning, Love2004SUSTAIN:Learning} so that they can apply to photographic stimuli processed by deep convolutional neural networks (DCNN). We evaluate basic hypotheses concerning the predicted performance benefits and costs associated with goal-directed, selective attention, which we view as distinct from bottom-up or saliency-driven capture \citep{Connor2004VisualTop-down,
Itti2001ComputationalAttention, Katsuki2014Bottom-upSystems}.

Attention-related approaches have met with great success in machine learning in key applications, such as machine translation \citep{Vaswani2017AttentionNeed} and image recognition \citep{Hu2018Squeeze-and-ExcitationNetworks}. Although motivated by attention in people, attention in machine learning often misses a critical component. Human attention is not just captured by the current \emph{bottom-up} context (e.g., a word or an image) but can also be driven by the current goal or expectations. Goal-directed attention is conspicuously absent in most DCNNs. Inspired by cognitive science and neuroscience research, our work aims to bridge this gap by offering a simple plug-and-play attentional mechanism that can be incorporated into pre-trained DCNNs. Our use of existing networks follows the intuition that the basic organisation of the visual system does not change, while one's goal does. Instead, the visual system is modulated by top-down, attentional signals.

In this work, we focus on the costs and benefits of goal-directed attention \citep[cf.][]{Plebanek2017CostsMiss}. When searching for one's keys, goal-directed attention benefits an agent by prioritising objects with key-like features, resulting in more efficient search. Goal-directed attention also exacts a cost. For example, key-like features of non-key objects will be amplified, increasing the likelihood of a false alarm. For example, when albedo is highly attended, a person may mistake shiny objects for a key. On the other hand, without goal-directed attention, the search process may be inefficient. We hypothesise that the intensity of goal-directed attention will alter the bias and sensitivity of a model, which will determine whether goal-directed attention results in a net benefit for the agent. With the correct amount of goal-directed attention, a model may successfully balance the costs and benefits such that sensitivity ($d^\prime$ in signal detection terms) is increased.

\begin{figure}
    \centering
    \includegraphics[scale=0.24]{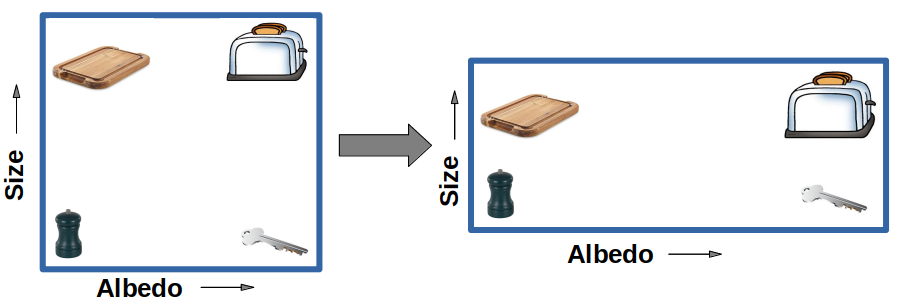}
    \caption{\textbf{Attention alters the importance of feature dimensions.} Four kitchen objects vary on two feature dimensions: \emph{albedo} and \emph{size}. In this example, albedo is the attended dimension (hence stretched) whereas attention to size is tuned down (hence compressed). Consequently, the key becomes more similar to the silver toaster than to the chopping board or salt shaker. Figure adapted from \citet{Nosofsky1986AttentionRelationship.}.}
    \label{fig:attention_warping}
\end{figure}

Psychologists and neuroscientists have developed models that include goal-directed attention to explain behavioural  \citep{Bar2006ARecognition, Itti1998ShortAnalysis, Love2004SUSTAIN:Learning, Miller2001AnFunction, Nosofsky1986AttentionRelationship., Plebanek2017CostsMiss, Wolfe1994GuidedSearch} and neuroimaging data \citep{Ahlheim2018EstimatingRepresentations, Mack2016DynamicKnowledge, Mack2020VentromedialLearning}. Algorithmically, goal-directed attention is often modelled as a set of weights that alters the importance of different psychological feature dimensions. Geometrically, one can think of attention weights as expanding and contracting the different feature dimensions of psychological space such that the goal-relevant dimension becomes more discriminative (Fig. \ref{fig:attention_warping}) \citep{ Kruschke1992ALCOVE:Learning, Love2004SUSTAIN:Learning, Nosofsky1986AttentionRelationship.}. 

Although the principles in these models are illuminating, the models are not directly applicable to deep learning as cognitive science researchers typically rely on low-dimensional, hand-coded stimulus representations, as opposed to photographic stimuli (i.e., pixel-level input). We aim to address this limitation by incorporating goal-directed attention mechanisms into DCNNs.

Given the success of of DCNNs, our aim of augmenting them is sensible. DCNNs without goal-directed attention do a good job accounting for patterns of brain activity along the ventral visual stream in tasks that do not emphasise goal-directed attention \citep{Schrimpf2018Brain-Score:Brain-Like}. What is lacking in these models is goal-directed attention selection, which is modulated by brain areas such as PFC \citep{Miller2001AnFunction}. These regional interactions allow the human brain to retool previously acquired features for a novel task, rather than learn new features from scratch. Our hope is that making DCNNs more brain-like will help address cases in which networks fall short of human performance, such as their susceptibility to natural adversarial images \citep{Hendrycks2019NaturalExamples}. 

\begin{figure}
    \centering
    \includegraphics[scale=0.25]{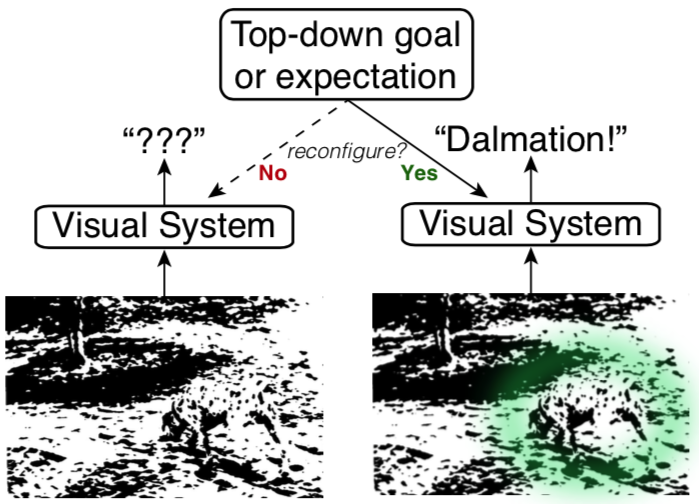}
    \caption{\textbf{Goal-directed attention signal.} Goal-directed attention is a top-down signal from outside the visual system that can reconfigure the visual system to reflect current goals and expectations. In this example, the absence of a strong top-down signal (left) to guide visual processing leads to uncertainty about what this confusing image depicts. In contrast, when there is an expectation that a dog is present (right) the visual system is reconfigured to be more sensitive and biased toward supporting information, which leads to successful recognition of the Dalmatian.
    }
    \label{fig:topdown3}
\end{figure}

Using DCNNs as a starting point, we incorporate a simple goal-directed attention mechanism motivated by research in psychology and neuroscience \citep{Nosofsky1986AttentionRelationship., Kruschke1992ALCOVE:Learning, Love2004SUSTAIN:Learning, Mack2016DynamicKnowledge, Mack2020VentromedialLearning}.
We cast goal-directed attention as driven by a top-down signal that is, in ways, separate from the visual system (Fig. \ref{fig:topdown3}). For example, the goal of locating one's keys does not primarily arise from activity (including recurrent activity) along the ventral visual stream. Instead, a higher-level goal, such as be on time for a morning meeting, leads to pursuing the subgoal of finding one's keys, which in turn engages the visual system in this search task. Our basic proposal differs from other attention strategies which lack a clear goal-directed component (e.g., \citealt{Bahdanau2015NeuralTranslate, Hu2018Squeeze-and-ExcitationNetworks}). 

We propose that an external expectation signal gives rise to a set of attention weights that specialises the visual system for the current goal. We implement and test this proposal by calculating a set of attention weights for each possible goal (e.g., is a cat present in this image). Our attention mechanism is implemented as a single trainable layer inserted into the mid-level of a pre-trained DCNN, permitting plug-and-play use. In accord with work in cognitive modelling and neuroscience, this feature-based attentional mechanism amplifies or attenuates activations within the DCNN depending on their goal relevance. At a very general level, our approach is in accord with neuroscience findings, such as increased or decreased firing rates for neurons as a function of whether the feature is attended \citep{Treue1999Feature-basedCortex} and warping of representational spaces (Fig. \ref{fig:attention_warping}) as revealed by the BOLD response \citep{Folstein2013CategoryCortex, Mack2016DynamicKnowledge}. 

Placing the attention mechanism at a mid-level, is in accord with  neuroimaging  results  demonstrating  that  the dimensionality  of  the BOLD response in lateral occipital cortex (LOC) varies with the number of task-relevant feature dimensions that are attended \mbox{\citep{Ahlheim2018EstimatingRepresentations}}, as well as recent work linking individual differences  in  selective  attention  (assessed  through  a  fit  of  a  cognitive model  to  behaviour)  to  patterns  of  brain  activity \mbox{\citep{Braunlich2019OccipitotemporalKnowledge}}. These results prompted us to evaluate whether mid-level attention modulation could be effective in a DCNN, as opposed to simply retraining the most advanced network layer to meet the task demands. In evaluating the attentional mechanism, we aimed to advance the range of cognitive processes that DCNNs can address in a brain consistent manner.

We aim for our approach to help explain basic psychological phenomenon, like how identification of an object in an image can be facilitated by a valid cue. For example, many people cannot identify any object in the photograph \citep{James1965SightEyes} shown in Fig. \ref{fig:topdown3}. However, when told there is a dalmatian in the photograph, the dog is immediately discernible. As a quick demonstration, we applied the same DCNN used in the studies reported here to the photograph. Like most people, the DCNN could not offer a clear interpretation of the photograph. Its top prediction was ``shower curtain'' with 5\% probability. In contrast, after giving the model an external signal to look for dalmatian (via appropriate goal-directed attention weights), the model predicted ``dalmatian'' as the most likely class with 55\% probability. 


Although compelling, rather than relying on anecdotal demonstrations, we systematically evaluated our top-down, goal-directed, selective attention mechanism by incorporating it into a popular DCNN \citep{Simonyan2015VeryRecognition} that was pre-trained on ImageNet \citep{DengJ.andDongW.andSocherR.andLiL.-J.andLiK.andFei-Fei2009ImageNet:Database}. ImageNet is a large-scale dataset of naturalistic images drawn from 1000 categories that can be divided into training\footnote{There are roughly 1,300 images per category.} and test images for cross-validation\footnote{There are roughly 50 images per category.} purposes.

\begin{figure}
    \centering
    \includegraphics[scale=0.24]{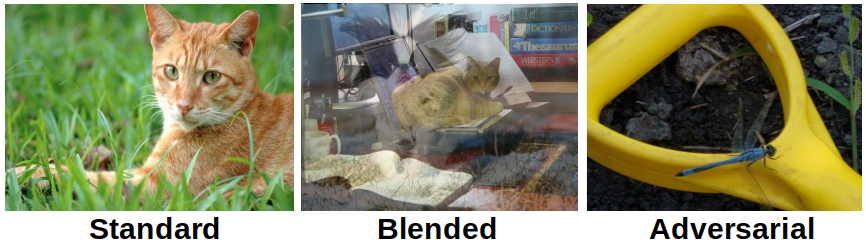}
    \caption{\textbf{Example stimuli from three categorisation problems.} (Left) A standard image used in Experiment 1 from ImageNet's Tabby Cat category \citep{DengJ.andDongW.andSocherR.andLiL.-J.andLiK.andFei-Fei2009ImageNet:Database}. (Middle) A blended image used in Experiment 2 made by alpha-blending an image of a cat and an image of a dog. (Right) A natural adversarial image used in Experiment 3 of a dragonfly missclassified as banana by DenseNet-121 with high confidence \citep{Hendrycks2019NaturalExamples}.}
    \label{fig:eg}
\end{figure}

Our attention mechanism was also trained using that same collection of photographic images from ImageNet's training set. We tested the attention mechanism in three simulation studies that involved progressively challenging image classification tasks (See Fig. \ref{fig:eg} for some examples). In Experiment 1, we evaluated our approach using \emph{standard} images from ImagNet's test set. In Experiment 2, we used \emph{blended} images, where each test image is created by alpha-blending two standard images. In Experiment 3, we used \emph{natural adversarial} images that are challenging for DCNNs to correctly classify \citep{Hendrycks2019NaturalExamples}. All three of our studies found that both the costs and the benefits of attention increased as goal-directed attention for the target increased, and that there was a net benefit of attention at moderate levels of target focus. The mid-level attention approach compared favourably to  retraining the final fully connected layer upon goal shifts.

\paragraph{Attention in Deep Learning} Attention approaches are gaining prominence in machine learning. We view self-attention as a form of bottom-up attention modulated by the current sequence of inputs rather than changes in goal-directed expectations \citep{Bahdanau2015NeuralTranslate, Chen2017SCA-CNN:Captioning, Vaswani2017AttentionNeed, Xu2015ShowAttention}. Self-attention is driven by the stimulus. For example, in machine translation, the contribution of each context word changes depending on the target word \cite{Bahdanau2015NeuralTranslate, Vaswani2017AttentionNeed}. To provide another example, in image classification, filters are amplified or suppressed based on the input image \citep{Hu2018Squeeze-and-ExcitationNetworks}. In contrast, our work learns goal-directed attention weights for different tasks. 
A second difference between our approach and self-attention is how attention weights are trained. Whereas self-attention is trained jointly with the rest of the network, we train the attention component separately from the rest of the network. Our attention mechanism is modular and designed to operate with any pre-trained DCNN. Unlike end-to-end fine-tuning which re-trains all parameters of a pre-trained DCNN for a new task \citep{Yosinski2014HowNetworks}, we only train the parameters associated with an attention layer. Our approach is motivated by the idea that prefrontal cortex (PFC) reconfigures existing networks to suit the current goal.

Although the notion of attention, particularly self-attention, is popular in deep learning (see \citet{Jetley2018LearnAttention} for a discussion), the distinction between bottom-up and goal-directed attention is rarely made \citep{Bahdanau2015NeuralTranslate, Chen2017SCA-CNN:Captioning,Hu2018Squeeze-and-ExcitationNetworks, Vaswani2017AttentionNeed}. We hope our work clarifies this distinction and is complementary to self-attention by addressing the neglected goal-directed component of attention.

Besides self-attention, the term goal-directed top-down attention is often coupled with the notion of recurrent or iterative processing. In our definition, a goal-directed signal must be originated from outside the system, which is different from the following work. In \citet{Wang2014AttentionalFeedback}'s attentional neural network, a top-down biasing signal is iteratively updated to facilitate feature selection. This iterative process involves consulting the input stimulus whereas in our work, the top-down signal is input independent. Similarly, \citet{Stollenga2014DeepConnections} adopts a reinforcement learning framework to achieve feature selection through an iterative process. And yet, multiple passes have to be made over the same input stimulus to achieve the optimal feature weighting.

There are other lines of work touching on goal-directed tuning, which have important theoretical and engineering differences from ours. \citet{Miconi2016TheresTask}, \citet{Chikkerur2010WhatAttention} and \citet{Perez2018FiLM:Layer} studied feature-wise attention, but only our work considered training a single layer to facilitate easy network reuse. \citet{Perez2018FiLM:Layer} feature-wise transformation is the closest to our implementation and yet they did not define their feature-wise weighting as goal-directed attention and allowed the weightings to be negative, which is not in agreement with non-negative modulation from the neuroscience literature. \citet{Wang2014AttentionalFeedback} discusses that a wrong top-down bias could astray the model to make a false alarm. However, we investigate this issue systematically by evaluating the effect with respect to the intensity of goal-directed bias using signal detection framework. Moreover, a single layer attention mechanism in our work is more interpretable than their generative set-up because our attention layer has a direct correspondence to entire feature maps.

\citet{Cao2015LookNetworks, Zhang2018Top-DownBackprop} studied the top-down goal-directed effect in the context of object localisation. Using the target class as a prior, they eliminate neurons in the model that do not contribute to the target response, resulting in spatial attention. In both of their work, the weighting on neuron activations cannot go beyond one, which departs from neuroscientific findings that neural activities can be turned up. The top-down modulation in their work is adjusted to different target images. In other words, unlike our model that has a generic attention tuning for images of the same class, their network reconfiguration is specific to each input image.

\mbox{\citet{Thorat2019TheRecognition}} studied the impact of a cue-based feedback approach on visual processing systems via manipulating the neural and representational capacity of the network. Our  contribution and \mbox{\citet{Thorat2019TheRecognition}} are very different.  For example, we use a DCNN as a model of visual processing whereas \mbox{\citet{Thorat2019TheRecognition}} used a shallow fully connected network. DCNN use is important due to its proposed relation to the ventral visual stream and their performance qualities. We also explore the impact of attention at different intensity levels. Our main focus is the costs and benefits of top-down attention. Finally, our range of tasks and model evaluation is very different.

In many ways, our work is most similar to \citet{Lindsay2018HowModel} in that they are informed by a neuroscience perspective and incorporate goal-directed attention into an existing DCNN. However, there are some key differences between these two efforts. Our key intuition is that the basic representations of the visual system are task-general and only reconfigured when receiving a top-down goal. The structure of the original network should not be permanently altered, much like how the human brain does not radical change structure after performing a task. In contrast, \mbox{\citet{Lindsay2018HowModel}} tuned each DCNNs as a binary classifier rather than preserving the original 1000-way output. Learning binary classifiers in this way is subject to fewer constraints on how the features should be reconfigured. This is more than a technical detail as we seek to parallel how we believe the brain deploys top-down attention, namely to reconfigure an existing network to tune it to the task at hand. Finally, one of our key aims is to trace out the costs and benefits of attention systematically as attention intensity varies. The way we incorporate attention strength into the loss term is in ways more straightforward and interpretable than the method used in \mbox{\citet{Lindsay2018HowModel}}.

\section{Methods}
In our approach, we add an attention layer to a pre-trained DCNN to modulate its activity according to the present goal, akin to how attention weights warp the representational space in cognitive models (Fig. \ref{fig:attention_warping}). As discussed below, we calculated a set of attention weights for each \emph{target class} (i.e., the current classification goal, such as detecting whether a cat is present in an image) at each \emph{attention intensity} level.

Our attention layer can be incorporated into any DCNN.  In this contribution, we used a pre-trained version of VGG-16 \citep{Simonyan2015VeryRecognition}. VGG-16 is a well-known, yet relatively simple architecture that scores well in benchmarks concerned with characterising both behaviour and brain response along the ventral visual stream \citep{Schrimpf2018Brain-Score:Brain-Like}. 

VGG-16 is a feed-forward DCNN model consisting of 23 layers with 138,357,544 trainable parameters. A subset of these layers can be grouped into five convolutional blocks. Each convolutional block consists of a series of convolution layers, pooling layer and non-linear activation function. The convolutional blocks are trained to extract and construct complex features from the raw input image. The last two layers of the network are fully connected layers that are trained to sort stimuli into 1,000 predefined categories based on features extracted from the preceding layers. 

VGG-16 was pre-trained on ImageNet \citep{DengJ.andDongW.andSocherR.andLiL.-J.andLiK.andFei-Fei2009ImageNet:Database}, a large-scale dataset of naturalistic images drawn from 1,000 categories based on the WordNet ontology \citep{Miller1995WordNet:English}. ImageNet is a popular benchmark in the computer vision community. As detailed below, we used ImageNet's training set (around 1.3 million images) to train our attention layer.

\begin{figure}
\centering
\includegraphics[scale=0.33]{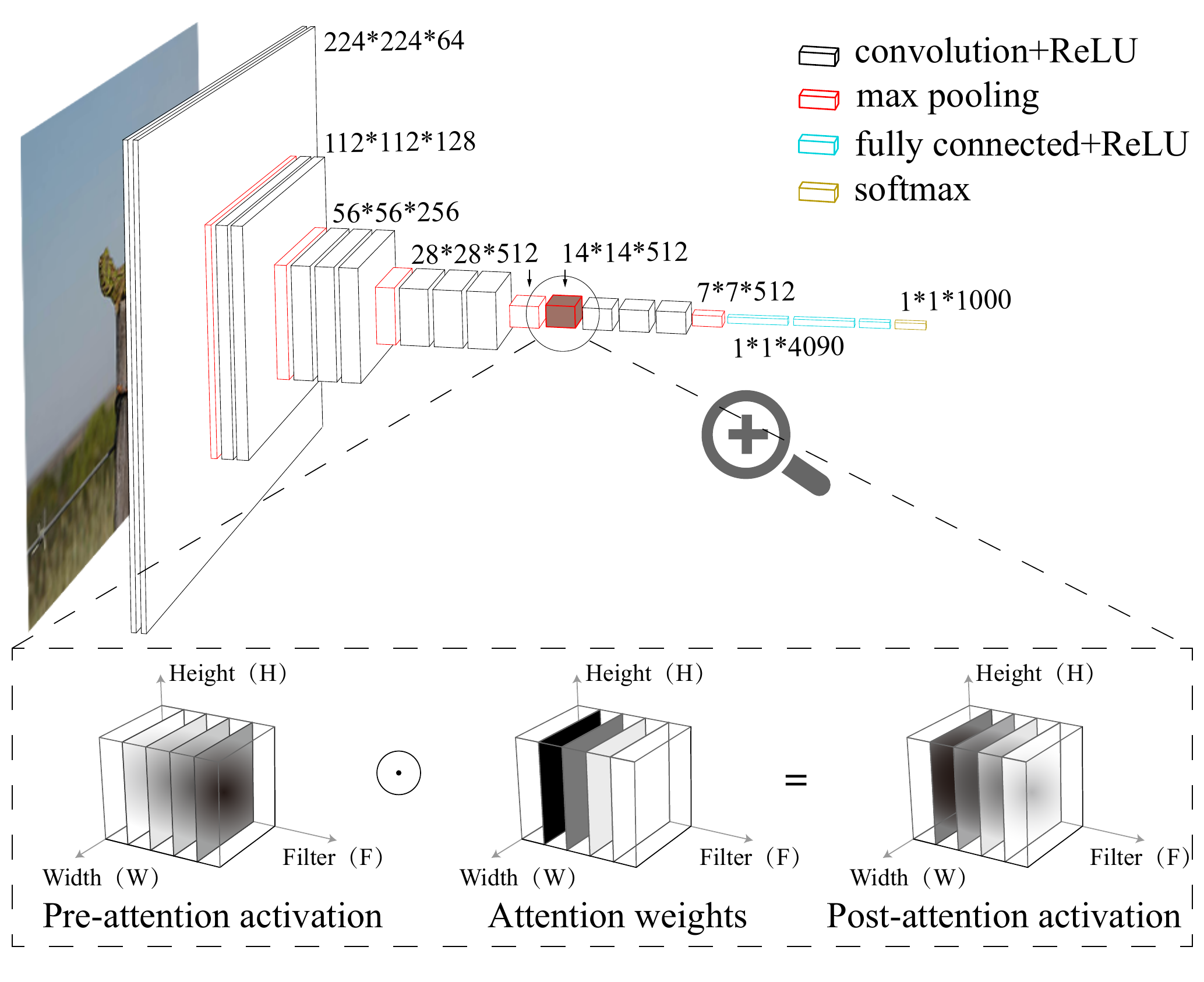}
\caption{\textbf{Integration of Attention Layer with VGG-16}. The top panel represents the architecture of VGG-16. An image will be processed from left to right and sorted into 1,000 categories. The layer preceding the attention layer (pre-attention activations) has an output representation with the shape of Height $\times$ Width $\times$ Filters. The attention layer is constructed with the same shape as the output representation but constrained such that a single filter value is used across all spatial locations. The attention operation is carried out as a Hadamard product between the pre-attention activations and attention weights. The post-attention activations have the same shape as the previous representation. The Hadamard product effectively re-weights the pre-attention activations using the corresponding attention weights. As the bottom panel shows, previously highly activated filter can be tuned down by a small attention weight (colour from dark to bright) whereas previously barely activated filter can become highly activated due to attention re-weighting (colour from bright to dark).}
\label{vgg+attention}
\end{figure}

\paragraph{Goal-directed Attention Layer}
A goal-directed attention layer can be inserted between any two layers of a pre-trained DCNN. Here, we inserted the goal-directed attention layer after VGG's fourth convolutional block, resulting in an attention layer of size 512, corresponding to the 512 filters of the preceding layer's output (see Fig. \ref{vgg+attention} for more details). The shape of the attention layer is equal to the shape of its preceding layer. The attention layer is connected in a one-to-one fashion to the filters of the preceding layer. These connections are referred to as \emph{attention weights} and modulate the activations of the preceding layer. We define this modulation as the Hadamard product (i.e., element-wise multiplication) between the preceding layer's activations and the attention weights.

In this work, we limit the flexibility of the attention layer by using a single weight to characterise all weights belonging to a particular filter, resulting in \emph{filter-wise} attention weights. Filter-wise attention weights embody the assumption that the attention weight for a particular filter should be spatially invariant. The attention weights are initialised to 1.0 and constrained to be between $[0, +\inf]$. During training, the attention weights are learned while the remaining network parameters are kept fixed, which we describe in more detail below.

\paragraph{Target Class}
A target class $\mathcal{T}$, is the set of all ImageNet classes that an attention network should focus on mastering. The remaining ImageNet classes are referred to as \emph{non-target classes}. In general, a target class can be composed of multiple ImageNet classes. In this work, a target class is composed of a single ImageNet class. A different target class implies a different task. 

\paragraph{Attention Intensity}
The attention intensity ($\alpha \in [0,1]$) determines the degree that the model should focus on mastering the target class at the expense of other non-target classes. 
Formally this is captured by weighting the contribution of each image to the loss term based on whether the image belongs to the target class or a non-target class (see the Model Training section). In this work, we consider five different intensity levels parameterised by $\alpha$ and $N$, where $N$ is the total number of classes in the categorisation task (Table \ref{alpha_table}). Since we use all ImageNet classes, $N$ is fixed at 1,000 throughout all studies.

When there is no selective attention, all $N$ classes are weighted equally, which means $\alpha=0.001$. This intensity level exhibits no goal-directed attention and primarily acts as a control model. There is very weak attention given to the target class when $\alpha=0.002$. When $\alpha=0.5$, the target class is weighted equally as all the non-target classes combined. When $\alpha=0.999$, the target class receives almost all the attention from the network. When $\alpha=1$, the network focuses exclusively on mastering the target class and non-target classes make no contribution to the loss term. We do not consider the case where $\alpha < 0.001$ because that would represent a model that does not attend to the target class which is beyond the scope of our studies.

Each level of intensity implies a trade-off between target and non-target performance. Re-weighting target and non-target classes can be likened to training on an imbalanced dataset. When $\alpha=1$, the learned attention weights will overfit to the target class given the absence of non-target classes. When $\alpha=0.5$, a balance is struck between target and non-target classes. We hypothesise that the largest net benefit will be achieved around this level. In addition to overall performance, the level of intensity should influence the distribution of learned attention weights. We expect that higher intensity levels will result in a larger number of filters being turned off.

\begin{table*}
\caption{\textbf{Sampled $\alpha$ values and meaning.}}
\label{alpha_table}
\begin{center}
\begin{small}

\begin{tabular}{lll}
\toprule
$\alpha$ ($N$: total number of classes) & Meaning & $\alpha$ ($N=1000$)\\
\midrule
$\frac{1}{N}$    & No selective attention & 0.001\\[.2cm]
$\frac{2}{N}$   & Weak attention & 0.002\\[.2cm]
$0.5$    & Balanced attention & 0.5\\[.2cm]
$\frac{N-1}{N}$    & Strong attention & 0.999\\[.2cm]
$1$   & Complete target focus & 1\\
\bottomrule
\end{tabular}
\end{small}
\end{center}
\end{table*}

\paragraph{Model Training}
The attention layer is trained using ImageNet-2012 \citep{DengJ.andDongW.andSocherR.andLiL.-J.andLiK.andFei-Fei2009ImageNet:Database}, the same dataset that was used to train the pre-trained network. 

In our work, we randomly sample 90\% of the training set images from each category as our training set. The remaining 10\% of the images are used for validation. We use the white-listed version of the validation set as our test set. We use the Adam optimiser \citep{Kingma2015Adam:Optimization} with a learning rate of 0.0003 and a batch size of 16. The training data are preprocessed and augmented according to \citet{Krizhevsky2012ImageNetNetworks} and \citet{Simonyan2015VeryRecognition}.

Given a target class ($\mathcal{T}$) and an attention intensity ($\alpha$), the attention weights are trained on a 1000-way classification problem, while keeping all other parameters fixed. The training objective is to minimise the standard multi-class cross-entropy loss (CE) over 1,000 classes. The contribution of each stimulus to the loss term is determined by whether it belongs to the target class. Stimuli that belong to the target class are weighted by $\alpha$.
Stimuli that do not belong to the target class are weighted $\frac{1-\alpha}{N-1}$. 

Formally, given a training image $x_i$ and its true class label $y_i$, the model outputs the probability for the true class $p(x_i)$. The $\alpha$-weighted cross-entropy $loss_i$ associated with image $x_i$ is defined in Equation (\ref{eq:loss_func}).

\begin{equation}
    loss_i = 
\begin{cases}
    \alpha CE(y_i, p(x_i)), & \text{if $x_i$ $\in$ $\mathcal{T}$}\\
   \frac{1-\alpha}{N-1} CE(y_i, p(x_i)), & \text{otherwise.}
\end{cases}
\label{eq:loss_func}
\end{equation}

Given that we train many attention models, it is computationally expensive to use the entire training set at each epoch. The computational cost is reduced by using a subset of the training set during each epoch. Each epoch uses all of the images belonging to the target class and a random 10\% of the images from each non-target class. The non-target samples change every epoch. The non-target samples are up-weighted in the loss term to adjust for the imbalanced sampling. The end result is that interpretation of attention intensity is unaffected by the sub-sampling procedure.

Models are trained using an early stopping criterion with a maximum of 5,000 training epochs. The early stopping criterion is based on the relative improvement in validation loss. Validation loss is computed using the weighted cross-entropy (Equation \ref{eq:loss_func}) and  all of images from the validation set. Relative improvement is computed between every other epoch. Training terminates when the relative improvement is less than 0.1\% at two consecutive checks. 

\section{Experiments}
The proposed goal-directed attention mechanism was evaluated using three experiments that used a shared training procedure. All experiments used the same set of trained models. One attention model was trained for each of the five attention intensities (Table \ref{alpha_table}) and each of 200 different target classes, yielding 1000 different attention models. Each target class corresponds to one of the classes defined in the natural adversarial dataset, a set of 200 classes which overlap with the ImageNet classes \citep{Hendrycks2019NaturalExamples}. 

Each experiment tested the trained models using a different image classification problem. Experiment 1 used \emph{standard} ImageNet images. Experiment 2 used \emph{blended} images that are made from two alpha-blended standard images. Experiment 3 used \emph{natural adversarial} images, taken from the natural adversarial dataset. 

Performance on each experiment was analysed using two distinct analyses. The first analysis examined how attention intensity affects the top-5 hit rate and top-5 false alarm rate of a testing model. Top-5 hit rate is simply a special case of hit rate where we consider the model makes a correct classification (i.e., true positive) so long as the correct class is among the top-5 predictions. Similarly, top-5 false alarm means we consider the model has made a false positive mistake when the class that the current model corresponds to (e.g., a Chihuahua attention model) appears in the top-5 predictions (e.g., predicting Chihuahua when there is no Chihuahua in the image).

The second analysis used signal detection theory to evaluate any changes in model sensitivity and criterion due to goal-directed attention \citep{MacMillan2002SignalTheory}. We expected a consistent pattern across all experiments such that increasing attention intensity would lead to higher benefits of goal-directed attention coupled with higher costs. The largest net benefit should be achieved when the target class and non-target classes are balanced ($\alpha=0.5$) or nearly balanced. We also expected that as attention intensity increases, more DCNN filters connected to the attention layer will be switched off (i.e., have an attention weight of 0). 

\subsection{Experiment 1: Standard Images}
Standard images from ImageNet are the most straightforward tests can be used to analyse the costs and benefits of the proposed goal-directed mechanism under normal conditions.

\paragraph{Testing Procedure}
Each trained attention model was tested using an equal number of target and non-target test images. All target class images and a random sample of non-target class images were used during testing. Using an equal number of target and non-target class images facilitates signal detection analyses. 

\begin{figure*}[ht]
    \centering
    \includegraphics[scale=0.45]{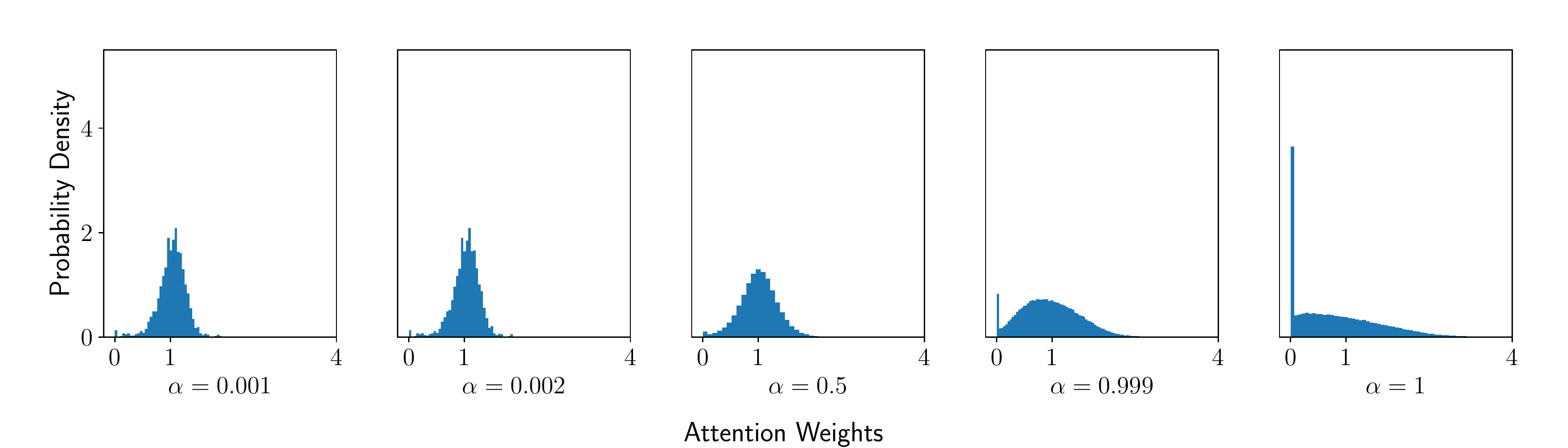}
    \caption{\textbf{Attention weight distributions for Experiment 1.}  As attention intensity increased, attention weights became more extreme (i.e., the variance of weights increased). Furthermore, increasing attention resulted in more filters being turned off (i.e., the initial attention weight goes from $1$ to $0$)}
    \label{fig:histogram}
\end{figure*}

\begin{figure*}
\centering
\includegraphics[scale=0.4]{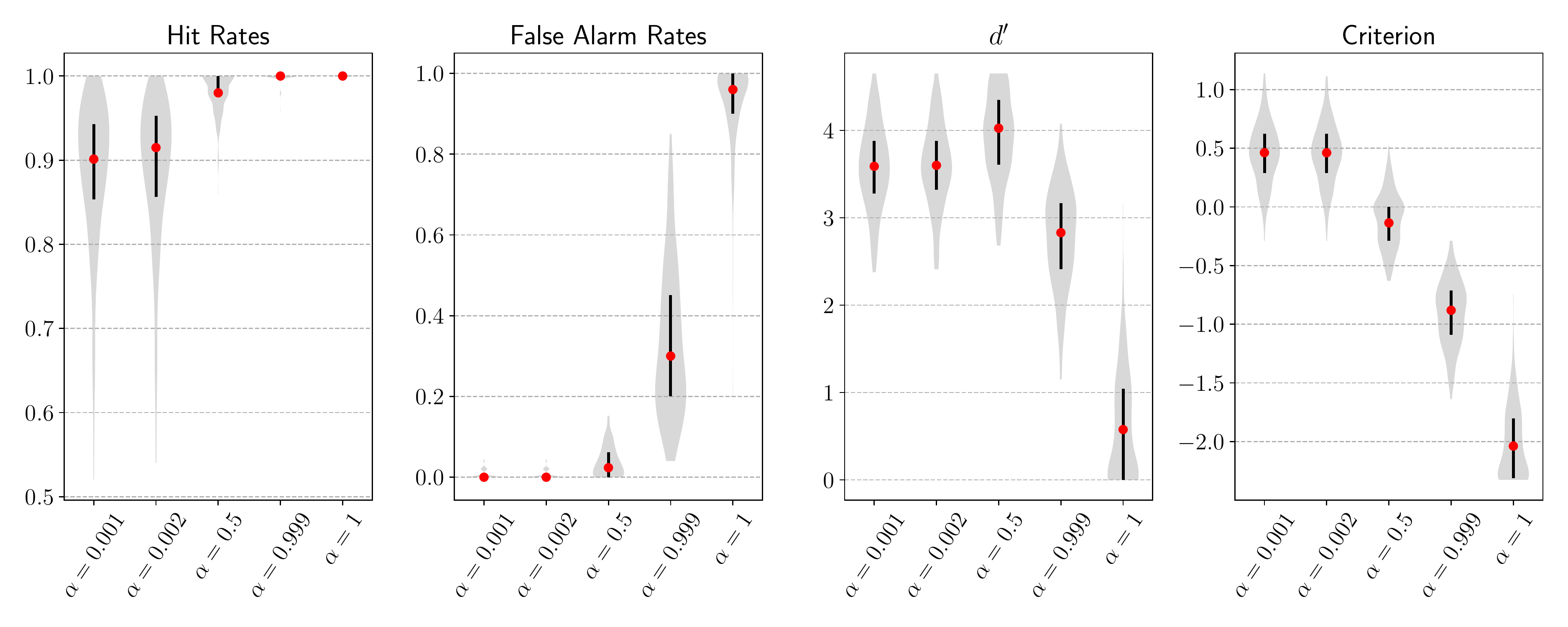}
\caption{\textbf{Main results from Experiment 1.} Goal-directed attention with varying degrees of intensity across all target classes were tested. As attention intensity increased, goal-directed attention had increasing benefits (higher model hit rates) as well as costs (higher model false alarm rates); With an increasing attention intensity, model sensitivity ($d^\prime$) first increased and then decreased. Model was more biased towards making a false alarm (criterion decreased). A sweet spot for maximal net benefit of goal-directed attention was achieved with the highest model sensitivity ($\alpha=0.5$). At this intensity, attention on the target class was equal to the attention on non-target classes combined.}
\label{fig:trg_non_trg}
\end{figure*}

\paragraph{Results}
Consistent with our hypothesis about the costs and benefits of attention, one might expect attention weights to become more extreme as attention was increased. Indeed, the variance of the attention weights increased as attention intensity increased and more filters were completely turned off due to zero attention weights (Fig. \ref{fig:histogram}). Although the system is nonlinear, the filters emphasised with increasing attention were largely consistent. Spearman correlations between adjacent $\alpha$ intensity values (0.001 to 0.002, 0.002 to 0.5, 0.5 to 0.999, 0.999 to 1.0) yield correlations of 0.99, 0.62, 0.63, and 0.87, respectively.

Both the hit rate and false alarm rate of the attention model increased as attention intensity increased (Fig. \ref{fig:trg_non_trg}). Model sensitivity ($d^\prime$) peaked near the balanced value of $\alpha=0.5$. Sensitivity difference across five attention intensities were significant, $F(4,995)=1285.4, p < .001$. The attention model with $\alpha=0.5$ had the highest sensitivity among sampled attention intensities and the difference to the second highest sensitivity ($\alpha=0.002$) was significant, $t(199)=10.1, p<.001.$ The model criterion monotonically decreased as $\alpha$ increased (Fig. \ref{fig:trg_non_trg}).

\paragraph{Discussion}
The results from Experiment 1 were consistent with our predictions in regards to the costs and benefits of attention. As attention intensity increased, attention weights became more extreme. Correspondingly, hits and false alarms both increased with increasing attention intensity. Sensitivity peaked near a moderate attention intensity that successfully balanced these benefits and costs. As expected, a sweet spot of attention intensity was shown around $\alpha=0.5$ from the sensitivity analysis when target and non-target classes were balanced. Decreasing model criterion suggests that as attention intensity increased, the model was more biased in favour of a target class response, which was more likely to result in a false alarm. Results on standard images demonstrated initial success of the proposed mechanism. Harder images with degraded features were used in Experiment 2 to further our understanding of the goal-directed attentional mechanism.

\subsection{Experiment 2: Blended Images}
Psychophysicists often use challenging visual tasks to probe important properties of the human visual system \citep{Yi2004NeuralLoad}. This experiment extended \citet{Lindsay2018HowModel}, which also used blended images to tax a goal-directed attentional mechanism. Blended images are harder tests for the model in that when two images are merged into one, features become overlaid and degraded.

\begin{figure*}
\centering
\includegraphics[scale=0.4]{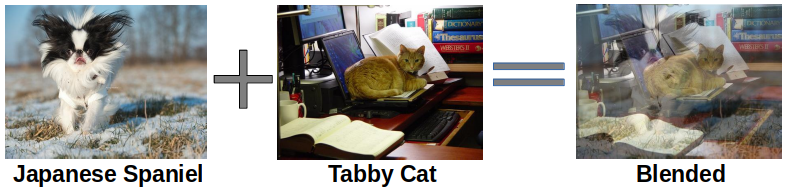}
\caption{\textbf{An example of how blended images are created.} Images from two categories are alpha-blended - each component image's pixel value is reduced by 50\% and added into one image.}
\label{blended_example}
\end{figure*}

\paragraph{Testing Procedure}
This experiment used the same attention models from Experiment 1, only the testing procedure differed. Blended images were created from the test set used in Experiment 1 (Fig. \ref{blended_example}) by combining images from two classes (e.g., Japanese Spaniel and Tabby Cat). There are two ground truth labels given one blended image. Upon testing an attention model for a given target class, it is considered a hit when the model correctly classifies the test image with the ground truth label that matches the target class of this model. For example, when we test a model with Japanese Spaniel attention, we consider the model makes a hit when it predicts Japanese Spaniel as the top-5 classes when Japanese Spaniel is one of the ground truth classes in the image. Similarly, we compute false alarm of an attention model by using blended images whose neither ground truth matches the target class of this model. For example, Japanese Spaniel attention model would make a false alarm when it predicts Japanese Spaniel as top-5 but there is no Japanese Spaniel in the image. For each testing model, we make sure there is an equal number of blended images that have and don't have the corresponding target class as its components. By doing so, we can further conduct signal detection analyses.

\paragraph{Results}
As attention intensity became stronger, the model hit rate and false alarm rate both increased. The sensitivity of the attention model increased at first and then decreased after balanced attention ($\alpha=0.5$). The overall sensitivity difference across the five levels of intensity was significant,  $F(4,198995)=10311.3, p<.001$. The highest net benefit was achieved when there was a moderate level of attention ($\alpha=0.5$) and was significantly higher than the next highest sensitivity ($\alpha=0.002$), $t(39799)=150.4, p<.001$. Additionally, model criterion dropped as attention intensity grew.

\paragraph{Discussion}
Classifying blended images is a more difficult problem than classifying standard images because the features of one class are superimposed on the features of another class. The difficulty of this experiment can be seen by comparing the results between the current experiment and Experiment 1 when no goal-directed attention was present ($\alpha=0.001$). The current experiment had a much lower baseline hit rate. It is a stronger demonstration that the proposed goal-directed mechanism was effective in selectively processing stimulus features in a task-specific manner. There is a consistent pattern to the previous experiment that increasing attention intensity improved the hit rate and false alarm rate, which suggests a clear trade-off between costs and benefits of attention. Consistent with our hypothesis that the largest net benefit was achieved when target and non-target classes were balanced ($\alpha=0.5$). As target and non-target classes becoming more imbalanced (increased $\alpha$), model criterion decreased, which indicates the model became more biased towards making a target class prediction over any test images.

\begin{figure*}
\centering
\includegraphics[scale=0.4]{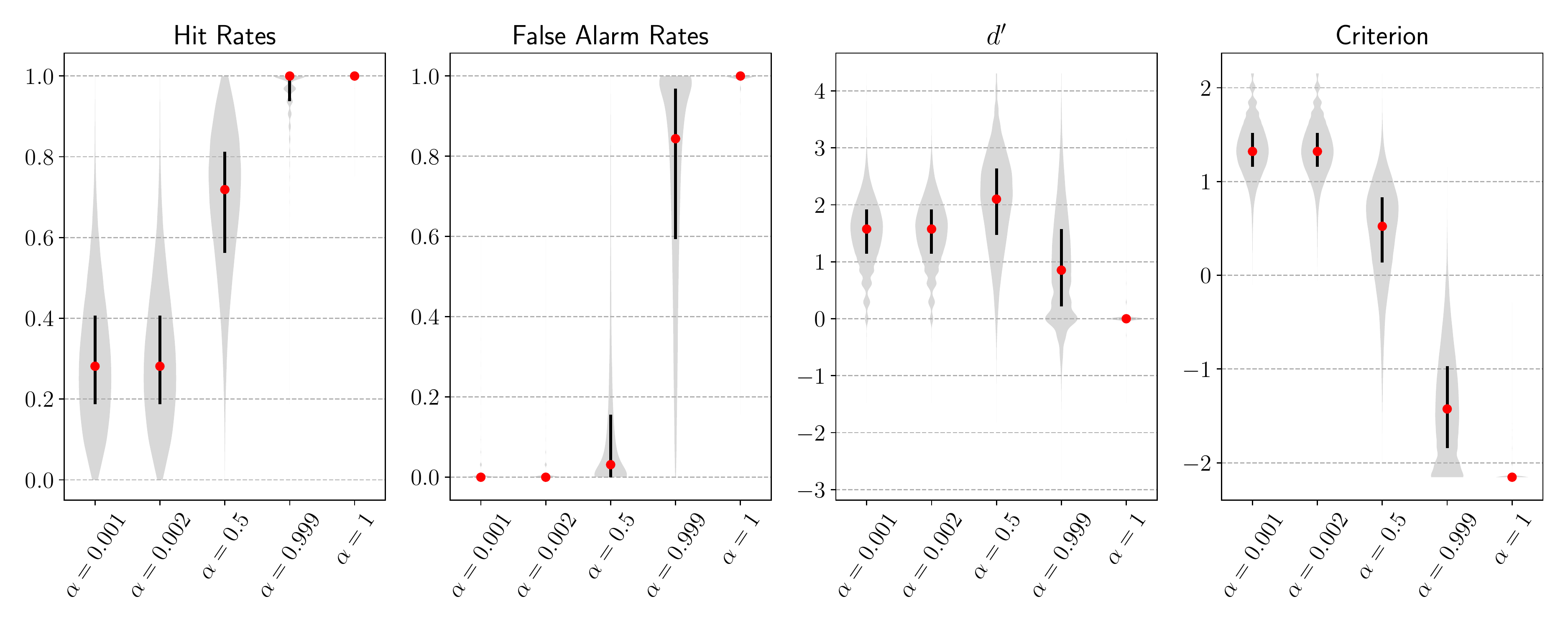}
\caption{\textbf{Main results from Experiment 2.} Goal-directed attention with varying degrees of intensity across all target classes were tested on blended images. Blended images were made of different classes of standard images that were hard for DCNNs to classify due to overlaid features. As attention intensity increased, goal-directed attention had increasing benefits (higher model hit rates) as well as costs (higher model false alarm rates); model sensitivity ($d^\prime$) first increased and then decreased. Model criterion was always decreasing, which suggests the model was increasingly biased to favour a target class prediction. A sweet spot for maximal net benefit of goal-directed attention was achieved when model sensitivity was the highest ($\alpha=0.5$), which was exactly the point where attention on the target class was equal to the attention on all non-target classes combined.}
\label{fig:exp2}
\end{figure*}

\subsection{Experiment 3: Natural Adversarial Images}
The final experiment uses natural adversarial images to evaluate the efficacy of goal-directed attention. The natural adversarial dataset is composed of 200 classes of real-world, unmanipulated images collected from the Internet \citep{Hendrycks2019NaturalExamples}. The classes have been intentionally selected to overlap with 200 classes from ImageNet. These images can reduce DCNN performance drastically by exploiting the vulnerabilities of these networks such as their colour and texture biases \citep{Guest2019LevelsCategorization, Hu2018Squeeze-and-ExcitationNetworks}. Although adversarial attacks have been heavily studied \citep{Goodfellow2015ExplainingExamples, Nguyen2015DeepImages, Song2018ConstructingModels}, these works use synthetic or unrealistic images that are carefully designed to defeat advanced DCNNs. Natural adversarial images offer an opportunity to test the proposed model with stimuli that humans would plausibly encounter in the environment.

\paragraph{Testing Procedure}
The same attention models were used, only the testing procedure differed. Each model was tested using an equal number of target and non-target images from the natural adversarial dataset. The same analyses were carried out on the test results.

\begin{figure*}
\centering
\includegraphics[scale=0.4]{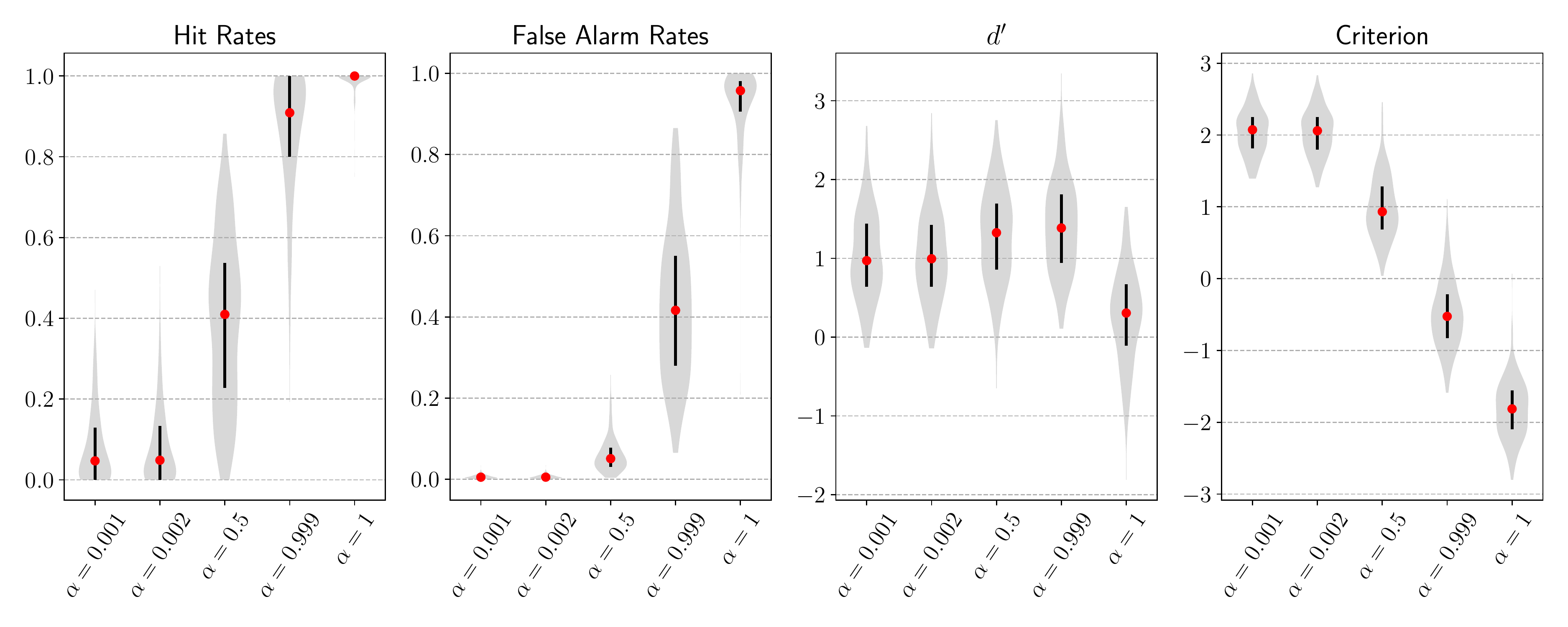}
\caption{\textbf{Main results from Experiment 3.} goal-directed attention with varying degrees of intensity across all target classes were tested on natural adversarial images. Natural adversarial images are unmodified natural images but challenging for DCNNs to classify correctly due to DCNNs' over-reliance on colour, texture as well as background cues. As attention intensity increased, goal-directed attention had increasing benefits (higher model hit rates) as well as costs (higher model false alarm rates); model sensitivity ($d^\prime$) first increased and then decreased. Model criterion decreased consistently, which means the model was becoming more likely to make a target class response regardless of images. A sweet spot for maximal net benefit of goal-directed attention was achieved when model sensitivity was the highest ($\alpha=0.999$).}
\label{fig:exp3}
\end{figure*}

\paragraph{Results}
The same pattern of results was observed (Fig. \ref{fig:exp3}) as in Experiments 1 and 2.  Increasing the attention intensity led to greater benefits (e.g., higher hit rate) and costs (e.g., higher false alarm rate). As in the previous studies, model sensitivity peaked for a moderate value of attention intensity. Model criterion decreased (biasing toward the target class) as intensity increased. The difference across models' sensitivities was significant, $F(4,995)=115.1, p<.001$. Model sensitivity was the greatest when $\alpha=0.999$, which was significantly higher than $\alpha=0.5$, $t(199)=3.5, p<.001$.

\paragraph{Discussion}
Like the previous two experiments, the proposed goal-directed mechanism achieved higher hit rates with higher false alarm rates as attention intensity increased. The optimal model sensitivity was found when attention intensity was $\alpha=0.999$. Model criterion shared the same pattern to previous studies, which suggests it was shifting to favour the target class responding as attention intensity becoming more extreme. There is a clear trade-off between costs and benefits of attention at different intensity levels. Unlike blended images, natural adversarial images exploit DCNNs' biases towards colour, texture and background cues. Although the goal-directed attention mechanism did not tackle these issues directly, the simple approach substantially improved performance. 

\section{A Non-Attentional Model Comparison}
The proposed attention mechanism has demonstrated its ability of reconfiguring a large network to better fit the current task goal. In machine learning, a different strategy of retraining the final network layer is commonly used to transfer knowledge from one task to a related task \citep{Yosinski2014HowNetworks}.

This alternative retraining approach is inconsistent with our theoretical aims. The mid-level attention mechanism is intended to reflect how the brain reconfigures select aspects of processing along the ventral visual stream. Attention effectively repurposes a large existing network through a relative small change (i.e., how the $512$ filters are weighted by attention). In contrast, even in our simulations, each task context requires $4,096,000$ specialty weights in the retraining approach, which raises scaling concerns.

Nevertheless, the retraining approach offers a useful yardstick against which to measure the performance of the attentional model. Given that retraining the final network layer is a popular and successful machine learning method for transfer learning, even nearing its performance levels would cast the attentional approach in a positive light.

To compare these two approaches, we used the same DCNN model for the retraining approach (absent the attention layer) and retrained the final layer using the same loss (i.e., goal) term that was used to train the attention weights. All the training and testing procedures were matched for both models. We compared the attention model and non-attention model at $\alpha=0.5$.

\begin{figure*}
    \centering
    \includegraphics[scale=0.4]{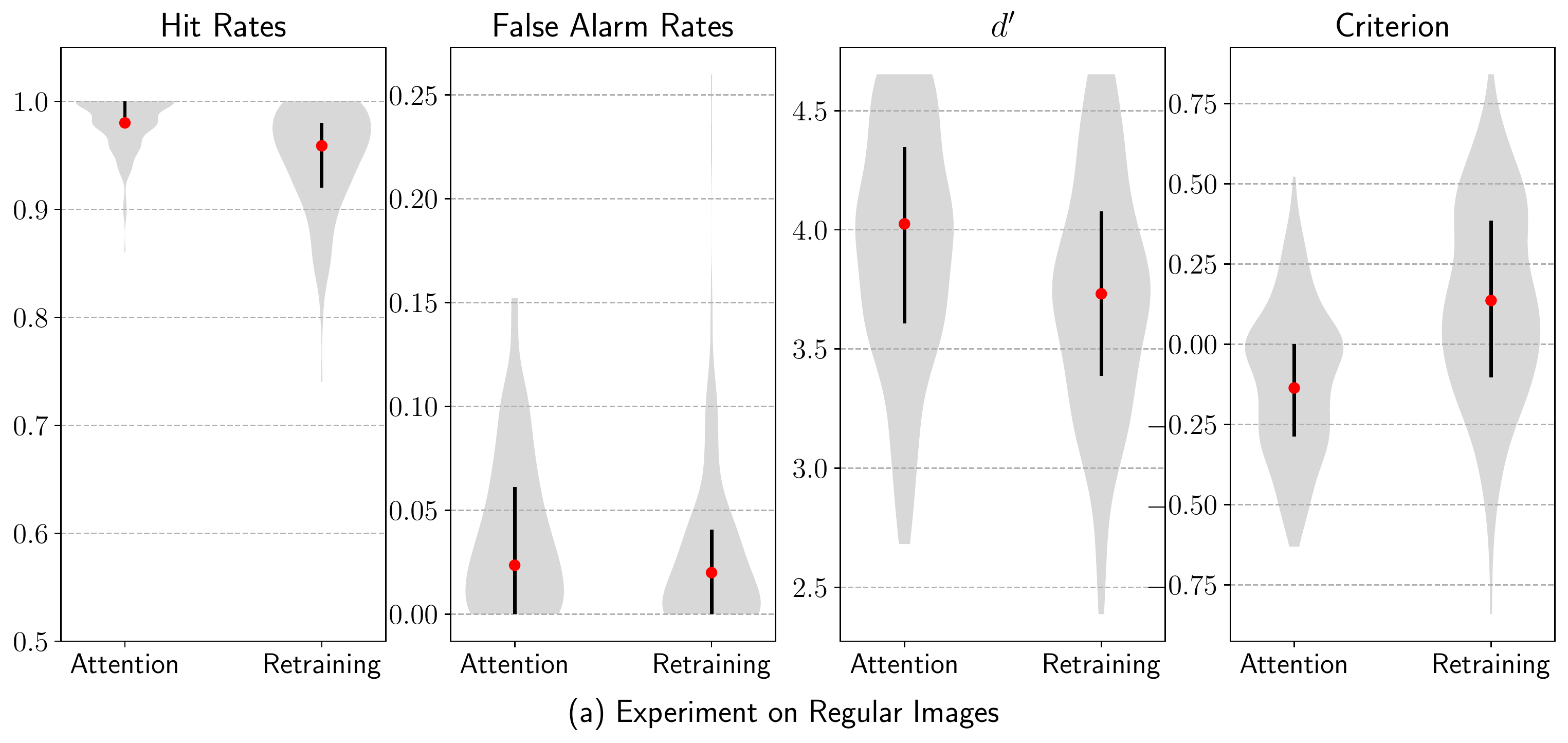}
    \includegraphics[scale=0.4]{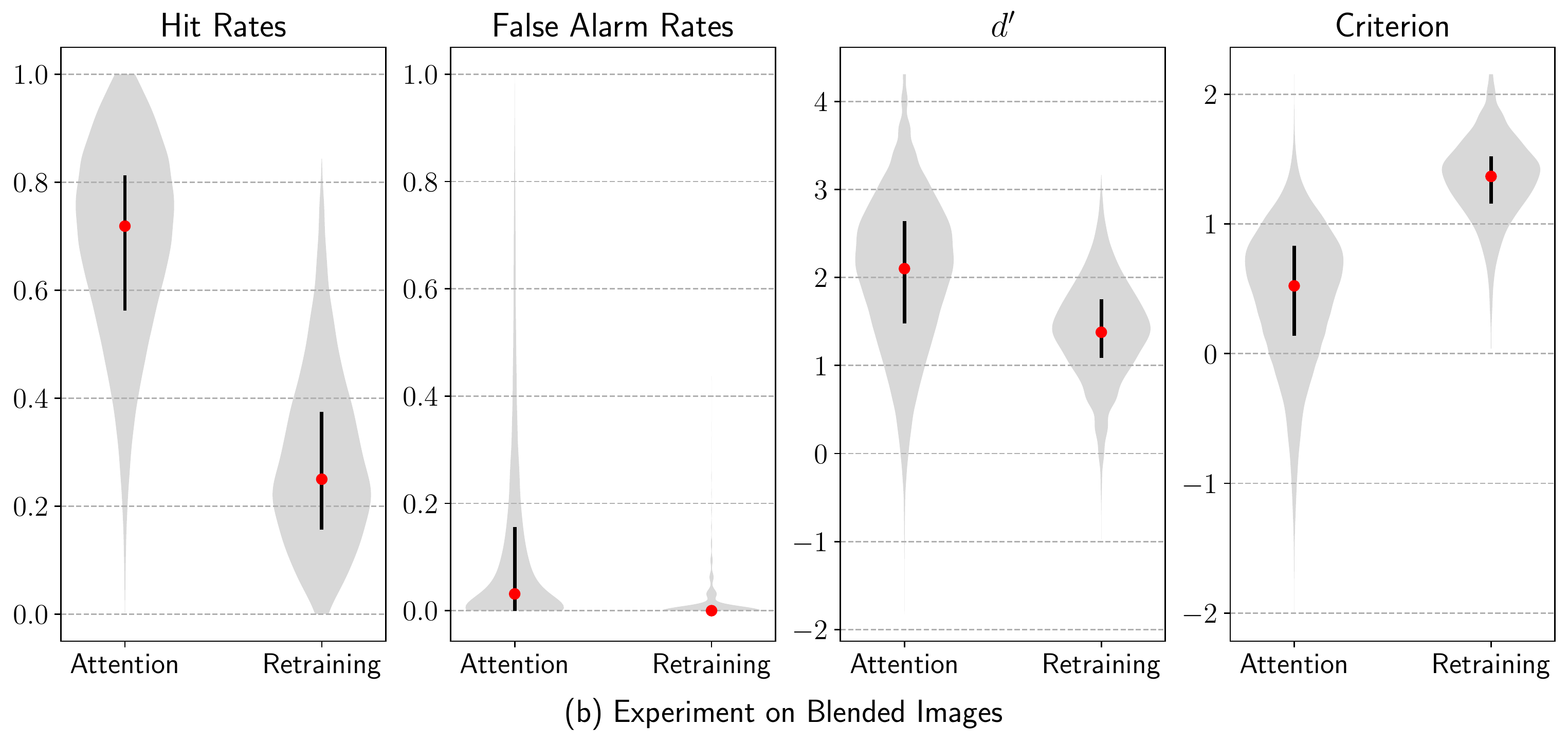}
    \includegraphics[scale=0.4]{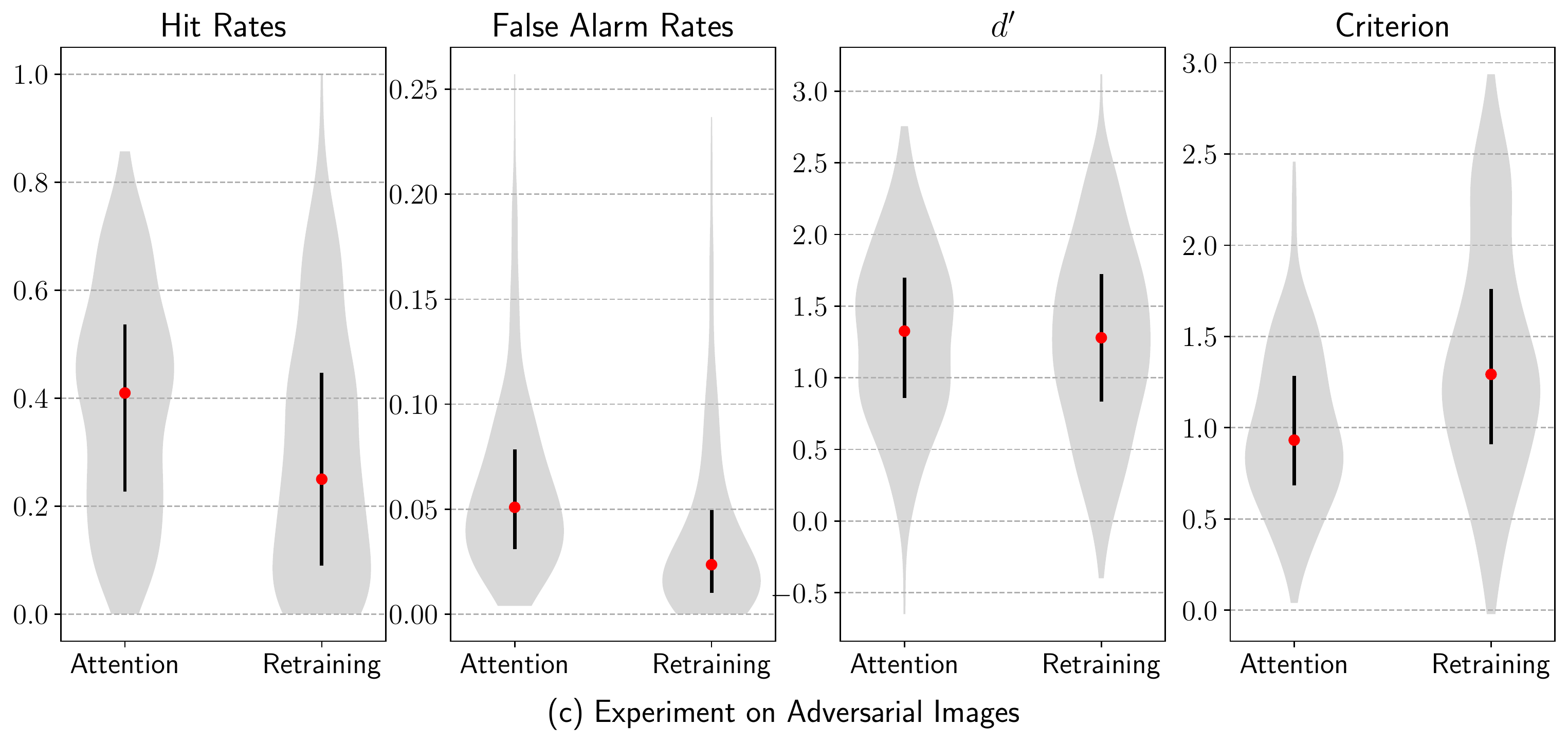}
    \caption{\textbf{Main results of comparing the Attention approach to the Non-attentional Retraining approach.} The proposed attention model and non-attention model trained at $\alpha=0.5$ were tested on regular images (a), blended images (b) and adversarial images (c). In the attention model, the attention layer is trainable. In the non-attention model, the attention layer is removed and the last fully connected layer is retrained according to the task goal. Both models were trained using the same procedures. Overall, the attention model performed better with a higher $d^\prime$ and lower criterion than the non-attention model.}
    \label{retraining_results}
\end{figure*}

\paragraph{Results}
The results across three experiments indicate a consistent performance boost when training an attention layer compared to training the last fully connected layer under the same training regime (Fig. \ref{retraining_results}). In particular, models trained with an attention layer result in a larger increase in hit rates relative to false alarm rates, resulting in a  higher $d^\prime$ measure.

\paragraph{Discussion}
Although the attentional model was conceived to advance and evaluate theory, surprisingly it performed better overall than a standard and successful approach to transfer learning from machine learning. One explanation for this success is that by the final layer it is too late for the model to modify stimulus representations to meet the task goal. In contrast, the attention layer interacts with subsequent network layers as the effects cascade through the network, which may provide additional flexibility to match the current task goal.

\section{General Discussion}

Motivated by research in psychology and neuroscience, we set out to test whether goal-directed attention could be successfully incorporated into pre-trained DCNNs for object recognition as a plug-and-play additional layer. Our aim was to extend attentional mechanisms found in category learning models from psychology to DCNNs that can take naturalistic stimuli as inputs. The theoretical idea evaluated was that goal-directed expectations (not driven by recent inputs) could reconfigure the existing network to specialise for the current task. In humans and non-human primates, this type of attention is thought to rely on goal-directed influences from prefrontal cortex. 

We evaluated some general hypotheses about how goal-directed attention should impact network performance. We predicted that as attention intensity (a hyperparameter in our model) increased, both the costs and benefits of attention should increase. We evaluated this hypothesis across three computational experiments, involving either standard images, blended images, or natural adversarial images. The basic prediction was that both the costs and benefits of attention should increase as attention intensity increases. We also predicted that there should be a sweet spot at moderate attention intensity where these benefits and costs would successfully balance.

All predictions held across all studies. We evaluated network performance in signal detection terms. Benefits, measured in terms of hit rate (e.g., responding Tabby Cat with Tabby Cat attention weights when a Tabby Cat is present), increased with increasing attention intensity. Likewise, costs, measured in terms of false alarm rate (e.g., responding Tabby Cat with Tabby Cat attention weights when a Tabby Cat is not present), increased with increasing attention intensity. Overall benefits, measured in terms of sensitivity (i.e., $d^\prime$), peaked for moderate levels of attention. We predicted the location of this peak as it was for an attention intensity setting that effectively balanced the importance of target and non-target items when training the attention weights. Bias toward the target category also increased with increasing attention intensity. Attention intensity can be viewed as a combination of the expectation and importance of detecting a member of the target category. It affects both bias and sensitivity, which seems consistent with human behaviour --- people have a tendency to both see what is expected and to perform better when focused appropriately.

Goal-directed attention appeared to reconfigure the network to specialise it for detecting the target class, much like how goal-directed attention reconfigures the human visual system when searching for a particular target (e.g., one's keys). Consistent with this notion, increasing attention intensity had the effect of increasing the variance of attention weights, which reweight filter responses, and turning off filters not relevant for detecting the target class (see Fig. \ref{fig:histogram}). One possible view is that the pre-trained DCNN effectively contains numerous subnetworks, many of which are not relevant to the current task and add unhelpful noise to the network response. Attention weighting could help by silencing irrelevant aspects of the network and amplifying relevant aspects. However, unlike hand-crafted features in cognitive models in which stimulus representations are intuitive, simple, and at a single level, it remains an open question whether filter representations in DCNNs correspond to our existing notions of psychological dimensions. In addition to considering behavioural measures, model representations can be evaluated against high-quality neuroimaging studies in which participants view naturalistic images (e.g., \mbox{\citet{Hebart2019THINGS:Images}}).

Although not a theoretical competitor, we compared the attention model's performance to a standard machine learning approach to transfer learning in which the final network layer is retrained for the current task. Unexpectedly, the attention model generally outperformed the machine learning approach. One explanation is that even though the attention layer had relatively few tunable parameters, the cascading effects through subsequent networks layers provided the needed flexibility to match the  task goal.

We set out to evaluate basic theoretical principles by evaluating attention weights trained for specific target classes. Although that satisfied our aims, successful systems and more encompassing models of humans may instead generalise across attention sets. For example, knowing what is relevant to attend when searching for a cat should overlap more with what is relevant when searching for a dog than for a truck. One solution is for the goal-directed signal itself to be a trained network that configures the attention weights for the current task goal. Such networks could also be endowed with the capability to search for conjunctions and disjunctions of target classes. We hope our results on target specific attention sets provide a solid foundation for future progress on this path.

Our goal-directed filter-based attention is distinct from work in spatial attention, though it could be related. For example, an attentional spotlight could move to areas of an image most responsible for driving goal-directed attention-weighted filters' responses. Likewise, our work could be extended to characterise the interplay of bottom-up saliency driven attentional capture with goal-directed goal-directed attention.

Neuroscience and machine learning have been enjoying a virtuous cycle in which advances in one field spurs advances in the other. For example, DCNNs were loosely inspired from the structure of the ventral visual stream and in turn have proven useful in understanding neuroscience data from these same brain regions. We hope that our work hastens this virtuous cycle by begetting more useful machine learning models that in turn inform our understanding of the brain.

\section*{Acknowledgements}
This work was supported by NIH Grant 1P01HD080679, Wellcome Trust Investigator Award WT106931MA, and Royal Society Wolf-son Fellowship 183029 to B.C.L.

\bibliography{main}
\bibliographystyle{icml2020}

\end{document}